\documentclass[letterpaper, 10 pt, journal, twoside]{IEEEtran}
\ifCLASSINFOpdf
\else
\fi
\usepackage{epsfig} 
\usepackage{mathptmx} 
\usepackage{times} 
\usepackage{amsmath} 
\usepackage{amssymb}  
\usepackage{graphicx}
\usepackage{booktabs}
\usepackage[noadjust]{cite}
\usepackage{orcidlink}

\begin{document}

\onecolumn

\begin{titlepage}

        \vspace*{1cm}
        
        {\normalsize\centering The following is the accepted version of the manuscript\\\vspace*{0.5cm}
        A. Ghafourian \textit{et al}., "Hierarchical End-to-End Autonomous Navigation Through Few-Shot Waypoint Detection," in \textit{IEEE Robotics and Automation Letters}, vol. 9, no. 4, pp. 3211-3218, April 2024, doi: 10.1109/LRA.2024.3365294.}
        
        \vspace{1.0cm}
        
        {\normalsize\centering Also appeared at the 40th Anniversary of the IEEE International Conference on Robotics and Automation (ICRA@40), 23-26 September, 2024, Rotterdam, The Netherlands.\\\textit{URL}: \centering\url{https://ras.papercept.net/conferences/conferences/ICRAX24/program/ICRAX24_ContentListWeb_3.html}.\\}
        
        \vspace{0.5cm}
        
        \vspace{17cm}
        
        {\small\noindent © 2024 IEEE. Personal use of this material is permitted.  Permission from IEEE must be obtained for all other uses, in any current or future media, including reprinting/republishing this material for advertising or promotional purposes, creating new collective works, for resale or redistribution to servers or lists, or reuse of any copyrighted component of this work in other works.}
        
        \vfill
        
\end{titlepage}

\twocolumn

%
\title{Hierarchical end-to-end autonomous navigation through few-shot waypoint detection}
%
%
%


\author{Amin Ghafourian$^{1 \orcidlink{0000-0003-3262-7775}}$, Zhongying CuiZhu$^{2 \orcidlink{0009-0000-2074-1445}}$, Debo Shi$^{3 \orcidlink{0000-0002-1327-7442}}$, Ian Chuang$^{4 \orcidlink{0000-0002-1983-9848}}$,\\Francois Charette$^{2 \orcidlink{0009-0002-8168-5904}}$, Rithik Sachdeva$^{3 \orcidlink{0009-0001-2379-3168}}$, and Iman Soltani$^{1 \orcidlink{0000-0001-9430-1522}}$
\thanks{Manuscript received: September 30, 2023; Revised December 16, 2023; Accepted January 22, 2024.}
\thanks{This paper was recommended for publication by Editor Aniket Bera upon evaluation of the Associate Editor and Reviewers' comments.
This work was supported by Ford Motor Company.}
\thanks{$^{1}$Amin Ghafourian and Iman Soltani (corresponding author) are with the Department of Mechanical and Aerospace Engineering, University of California, Davis, Davis, CA 95616, USA {\tt\small aghafourian@ucdavis.edu, isoltani@ucdavis.edu}}%
\thanks{$^{2}$Zhongying CuiZhu and Francois Charette are with Ford Motor Company, Palo Alto, CA 94304, USA
        {\tt\small cuizy3@gmail.com, charette.francois@outlook.com}}%
\thanks{$^{3}$Debo Shi and Rithik Sachdeva are with the Department of Electrical and Computer Engineering, University of California, Davis, Davis, CA 95616, USA {\tt\small deshi@ucdavis.edu, rithiksachdeva@gmail.com}}%
\thanks{$^{4}$Ian Chuang is with the Department of Computer Science, University of California, Davis, Davis, CA 95616, USA {\tt\small itchuang@ucdavis.edu}}%
\thanks{\textbf{Video attachment:} https://youtu.be/G4QQbESYeas}%
\thanks{\textbf{Dataset:} https://ucdavis.box.com/s/9tt9usez75uj5l551sgtghgpbgsuc955}%
\thanks{\textbf{Code:} https://github.com/Soltanilara/DNS.git}%
\thanks{Digital Object Identifier (DOI): 10.1109/LRA.2024.3365294}
}

\markboth{IEEE Robotics and Automation Letters. Preprint Version. Accepted February, 2024}
{Ghafourian \MakeLowercase{\textit{et al.}}: Hierarchical end-to-end autonomous navigation through few-shot waypoint detection} 

%



\maketitle

\begin{abstract}
Human navigation is facilitated through the association of actions with landmarks, tapping into our ability to recognize salient features in our environment. Consequently, navigational instructions for humans can be extremely concise, such as short verbal descriptions, indicating a small memory requirement and no reliance on complex and overly accurate navigation tools. Conversely, current autonomous navigation schemes rely on accurate positioning devices and algorithms as well as extensive streams of sensory data collected from the environment. Inspired by this human capability and motivated by the associated technological gap, in this work we propose a hierarchical end-to-end meta-learning scheme that enables a mobile robot to navigate in a previously unknown environment upon presentation of only a few sample images of a set of landmarks along with their corresponding high-level navigation actions. This dramatically simplifies the wayfinding process and enables easy adoption to new environments. For few-shot waypoint detection, we implement a metric-based few-shot learning technique through distribution embedding. Waypoint detection triggers the multi-task low-level maneuver controller module to execute the corresponding high-level navigation action. We demonstrate the effectiveness of the scheme using a small-scale autonomous vehicle on novel indoor navigation tasks in several previously unseen environments.
\end{abstract}

\begin{IEEEkeywords}
Vision-Based Navigation, Motion and Path Planning, Deep Learning for Visual Perception, Motion Control, Representation Learning
\end{IEEEkeywords}

%
\IEEEpeerreviewmaketitle

\section{Introduction}
%
%
%
%
\IEEEPARstart{A}{ccurate} positioning, such as through visual or LiDAR SLAM or GPS, is crucial for mobile robotics, but often comes with high computational and hardware costs~\cite{debeunne2020review,cheng2022review}. Challenges also arise in environments with signal obstruction, such as in urban areas with dense constructions or underground passages. Despite potential future advancements in sensing technologies, the need for alternative navigation tools remains, as redundancy and simplicity are key factors for widespread adoption of mobile robots and autonomous vehicles.

This paper introduces a novel approach to autonomous vehicle navigation, inspired by human-like navigation using visual landmarks and simple instructions. Our system, termed Description-based Navigation System (DNS), simplifies navigation by associating limited visual data of key waypoints with high-level navigation actions (e.g., turning instructions). DNS minimizes reliance on complex localization sensors, leveraging a few-shot learning technique for waypoint detection. This approach is practical given the public accessibility of visual data from sources like Google Street View.

Our contributions include: 1) A hierarchical end-to-end navigation system separating low-level maneuvering from high-level navigation; 2) An efficient few-shot learning method for robust waypoint detection; 3) Demonstration through experiments of how well the waypoint detection integrates with conditional maneuver control in various settings.
\begin{figure}[t!]
    \centering
    \includegraphics[width=1.0\linewidth]{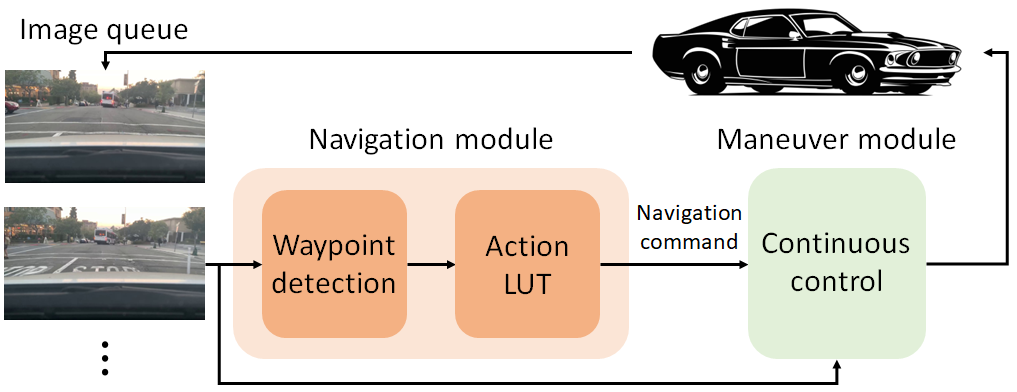}
    \caption{Proposed workflow. One or a few example images are used by the mobile robot to detect predefined landmarks in the environment. Upon landmark detection, the corresponding high-level discrete navigation action (e.g. turn right, turn left, etc.) is retrieved from a lookup table and passed to a continuous maneuver controller. Continuous control executes the resulting high-level navigation action while avoiding obstacles and maintaining the vehicle on a drivable path.}
    \label{fig:workflow}
\end{figure}

The paper hereafter is organized as follows: in Section~\ref{sec:relatedwork} we review some of the prominent related literature. Section~\ref{sec:memnav} discusses the proposed description-based modular navigation, and in Section~\ref{sec:fsl} the proposed few-shot technique and its deployment in the context of DNS is explained. In Section~\ref{sec:exp} we demonstrate the performance results for offline as well as real-time implementation of DNS and present the associated ablation study. Section~\ref{sec:conclusion} concludes the paper and presents future opportunities.

\section{Related Works}\label{sec:relatedwork}
In recent years, localization, path planning, and motion control for autonomous navigation have been active areas of research. Various sensors including cameras, GNSS (Global Navigation Satellite System) receivers, IMU (Inertial Measurement Unit) devices, visual and thermal cameras, and LiDAR sensors have been adopted to develop autonomous functions using various AI technologies as well as classical control paradigms. In this section, we review some of the recent works in this domain.

\cite{nilwong2019deep} and \cite{foroughi2021cnn} propose CNN-based outdoor and indoor localization techniques based on landmark detection and image classification. Both methods importantly require training on labeled data from the target environment. Further, the proposed technique in~\cite{nilwong2019deep} requires geolocation labels including coordinates and compass orientation, as well as bounding box annotations. In~\cite{choi2019map}, GNNS/INS (Inertial Navigation System) is used for highway localization, which is further refined against detected signs and road facilities. In~\cite{ort2022autonomous,ort2022maplite}, the need for HD prior maps for localization and subsequent planning is relaxed by using Standard Definition (SD) maps paired with onboard visual and LiDAR perception for inferring the HD map online. Further, precise localization is obtained using a Localizing Ground Penetrating Radar (LGPR) to retrieve stable underground features that are robust to weather changes~\cite{ort2022autonomous}. \cite{li2021openstreetmap} combines OpenStreetMap road network with GPS and IMU signals, as well as local perception information obtained by 3D-LiDAR and CCD camera sensor fusion to refine the global localization, as well as local path planning and control of the robot.

For local planning and motion control,~\cite{panda2023agronav} uses semantic line detection and segmentation to identify traversable lanes for agricultural robots and vehicles, tailoring their design to the order and configuration of the scene in that application. \cite{miyamoto2019vision} also uses image segmentation paired with topological maps for road following and control. They propose a rule-based heuristic design for explicitly controlling the robot toward a target point obtained through extrapolating and intersecting the road boundary lines and accounting for obstacles and drivable areas. \cite{dall2021fast} emphasizes performance in seen environments and route repeating capability through teach and repeat that utilizes odometry information and generates correction signals through lightweight processing of the visual input. For improving autonomous motion control,~\cite{liu2020driving} proposes incorporating prediction of road drivers' intentions through hidden Markov models. Some techniques are also specifically designed to perform specialized maneuvers such as lane change~\cite{wu2020trajectory} and overtaking~\cite{ortega2020overtaking}.

Imitation learning~\cite{couto2023hierarchical,karnan2022voila} and reinforcement learning~\cite{sutera2020indoor,dukkipati2022learning,wasala2020trajectory,wang2023inverse,xu2022learning} have also been attractive domains in mobile robotics research. \cite{couto2023hierarchical} generates a BEV representation that also denotes the desired route given camera input and high-level navigation action. The navigation action is obtained using GPS route planning. The capability for conditional BEV generation is gained through adversarial training. Given the BEV prediction and the current state of the vehicle, control policy is learned from expert demonstrations. Contrary to our method, BEV generation is an essential facet of this work, without which the tests are shown to fail. In~\cite{karnan2022voila}, the authors demonstrate how by utilizing Imitation from Observation (IfO) a robot can learn a good navigation policy for a route using the demonstrator's ego-centric video despite viewpoint mismatch. The policy can generalize to unseen environments if a recording is provided.
In our work, we primarily teach the route to the robot by presenting only minimal waypoints en route along with their corresponding high-level navigation actions without particularly focusing on the detailed low-level control. In VOILA, however, the main focus is the imitation quality and learning a policy that adheres to the demonstrated path from a differing platform at a low level without emphasizing waypoint detection and executing correct high-level navigation actions in particular.

The past few years have been a fertile period for high-performing vision models that are especially suited for use as backbone or in applications involving image retrieval. Notably, contrastive and non-contrastive joint-embedding self-supervised learning (SSL) techniques~\cite{chen2020simple,caron2020unsupervised,grill2020bootstrap,chen2021exploring,bardes2021vicreg} have provided effective means of leveraging unlabeled data to train robust models for various downstream applications. In these methods, the aim is predominantly to make representations robust against changes to the input data that do not change its semantics (e.g.~common visual transforms). With the increased availability of labels on large datasets, supervised counterparts such as~\cite{khosla2020supervised} trained on a similar objective but with real labels added to the surrogate labels have also emerged. An especially attractive model for universal visual place recognition was presented in~\cite{keetha2023anyloc}, which aggregates per-pixel features from self-supervised pretrained vision transformers (ViT)~\cite{caron2021emerging,oquab2023dinov2}. Such a model can facilitate image retrieval and localization for robot navigation.

In the following section, we discuss the Description-based Navigation System (DNS), a control scheme that places special emphasis on data efficiency and quick adaptation to unseen routes without fine-tuning.  

\section{Description-Based Navigation System}\label{sec:memnav}
The proposed process of setting up a mobile robot for an upcoming route consists of the following steps:

\begin{enumerate}

    \item Identify a set of waypoints along the route at locations where a new high-level navigation action should be executed, e.g. at an intersection where a vehicle should take a turn.
    
    \item Assign a discrete navigation action to each waypoint and form an address lookup table (LUT).
    
    \item Retrieve example images of each waypoint location: In this approach, we are assuming that for each waypoint, one or a few example images are available.
    
    \item Provide the waypoint representations to the vehicle so that they become identifiable upon future exposure to similar visual cues.
    
\end{enumerate}

\begin{figure}[t!]
    \centering
    \includegraphics[width=0.9\linewidth]{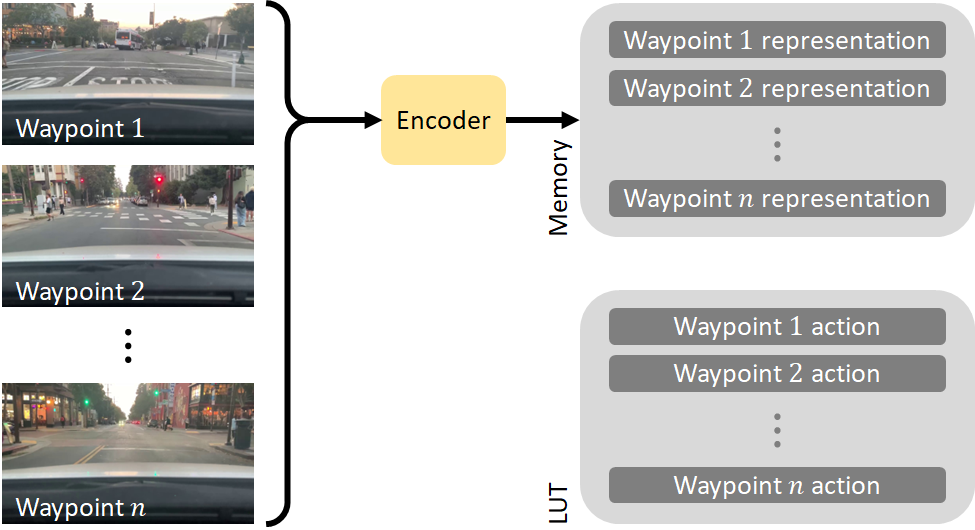}
    \caption{Route teaching stage. Prior to vehicle departure, the image representations from predetermined waypoints are used to populate the corresponding memory slots along with the high-level navigation action for future reference.}
    \label{fig:teaching}
\end{figure}

Given the limited number of samples available for each landmark as well as the variety of potential routes in many applications, few-shot learning is adopted to train the high-level navigation model so that it can quickly adapt to previously unseen waypoints. Individual memory slots are assigned to specific landmarks and paired with their corresponding high-level navigation actions such as a right or a left turn, U-turn, etc. A minimal number of waypoints and their corresponding navigation actions are specified to uniquely guide the vehicle or mobile robot at, for instance, the street, corridor, or hallway level, while providing adequate information to the considered low-level maneuver control module. Memory slots are populated using the extracted landmark representation (see Section~\ref{sec:fsl} for more detail). The above process, hereafter referred to as the route teaching stage, is schematically shown in Fig.~\ref{fig:teaching}

\begin{figure}[t!]
    \centering
    \includegraphics[width=0.85\linewidth]{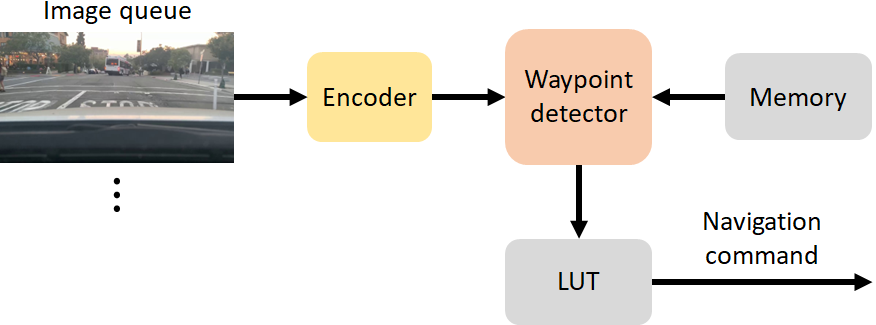}
    \caption{High-level navigation module. During inference, the navigation module processes the incoming images to compare them against memory content and detect waypoints. Upon detection, the corresponding high-level navigation action is retrieved from the lookup table and issued to the maneuver control unit.}
    \label{fig:navmodule}
\end{figure}

At drive time, the captured images are continuously processed using the model. The generated representations are compared against memory content associated with the upcoming waypoint for landmark detection. Upon recognition of a waypoint, the vehicle refers to LUT to retrieve the corresponding high-level action and feed it to the low-level maneuver control unit for execution. This process is schematically demonstrated in Fig.~\ref{fig:navmodule}.

\begin{figure}[t!]
    \centering
    \includegraphics[width=0.9\linewidth]{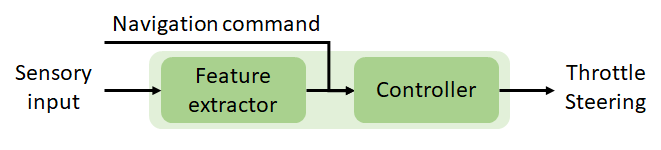}
    \caption{The low-level maneuver control module is composed of a feature extractor that processes the camera input, which is then concatenated with the discrete navigation action (e.g. turn right) and presented to the continuous controller to obtain steering and acceleration/deceleration outputs.}
    \label{fig:controller}
\end{figure}

The maneuver control unit is a neural network that receives incoming images paired with high-level action conditions and controls the vehicle to execute the desired action (Fig.~\ref{fig:controller}). The end-to-end maneuver control module is trained via imitation learning to receive images as input and, conditioned on the high-level action, issue acceleration, deceleration, and steering commands. This controller maintains its latest action condition until a different high-level decision associated with a newly detected waypoint is made.

\section{Few-shot learning for waypoint detection}\label{sec:fsl}

In this section, we describe the formulation of few-shot classification and explain how the proposed few-shot technique is adapted for waypoint detection.

\subsection{Few-shot classification: problem formulation}
As part of a few-shot classification training, we are given dataset $\mathcal D=\{({\bf x}_i,y_i)\}_{i=1}^N$ of labeled samples, where each $y_i\in\{1,\dots,K\}$ is the label of the corresponding example ${\bf x}_i$. For training, many $w$-way $s$-shot classification tasks, or episodes, can be randomly formed, where $w$ denotes the total number of participating classes and $s$ denotes the number of labeled examples or shots, commonly referred to as the support set, for each class. The model is then trained by minimizing classification loss on $q$ unlabeled samples (queries). Trained under this scheme, the model learns to rely on a small number of labeled samples (shots) for new tasks with novel participating classes.

\subsection{Enhanced metric few-shot learning}

In metric few-shot learning techniques, one aims to find a suitable representation that places instances from the same class close to each other while separating instances from different classes in the embedding space. This will facilitate accurate classification. In Prototypical Networks~\cite{snell2017prototypical}, for instance, this is done by minimizing the Euclidean distance between a query and its true class prototype defined as the mean of the support sample representations for that class, while maximizing its distance from all other class prototypes.

Deep variational embeddings in the context of few-shot learning have been shown to improve classical metric few-shot learning techniques~\cite{zhang2019variational,fort2017gaussian}, which can suffer from noise due to data scarcity and lack of interpretability. In these methods, rather than estimating a class mean based on a few sample vector representations, a more expressive class representation is adopted, for instance, by assuming a Gaussian form. In~\cite{zhang2019variational}, each sample is mapped to a Gaussian distribution of its associated class through a neural architecture, whose output is interpreted as mean and the variances that form a diagonal class covariance matrix. Individual sample outputs from each class are then combined to obtain a more accurate estimate of class mean and covariance.

Similarly, we retrieve class distribution mean and diagonal covariance estimates for each sample through a neural network and combine samples to form a more accurate class distribution. For class $k$, we obtain the combined distribution $\mathcal{S}^k=({\bf \mu}^k,{\Sigma}^k)$ by obtaining combined estimates for mean and covariance. The combined estimate of class $k$ mean is simply the average of individual mean estimates:
\begin{align}
    {\bf \mu}^k=\frac{1}{s}{\sum_{i=1}^{s}{{\bf \mu}_i}}.
    \label{eq:mean}
\end{align}

We adopt a linear combination of diagonal covariance estimates to obtain a combined minimum variance estimate~\cite{graybill1959combining}. Assuming equal weights, it reduces to an arithmetic mean:
\begin{align}
    {\Sigma}^k=\frac{1}{s}{\sum_{i=1}^{s}{{\Sigma}_i}}.
    \label{eq:cov}
\end{align}

Given a query ${\bf x}_j$, the associated distribution $\mathcal{Q}_j=({\bf \mu}_j,{\Sigma}_j)$ is obtained from the network and viewed as the estimate of the class distribution associated with that query. To quantify the distance between class support and queries we propose the use of a distribution-to-distribution symmetrized Mahalanobis distance. In this form, the distance between the query and class $k$ distribution becomes
\begin{align}
    d^2(\mathcal{Q}_j,\mathcal{S}^k) = ({\bf \mu}_j-{\bf \mu}^k)^T({\Sigma}_j^k)^{-1}({\bf \mu}_j-{\bf \mu}^k),
    \label{eq:mahal}
\end{align}
where
\begin{align}
    {\Sigma}_j^k=\frac{1}{2}\left({\Sigma}_j+{\Sigma}^k\right).
    \label{eq:symsigma}
\end{align}    

\subsection{DNS with distribution embeddings}

\begin{figure*}[t!]
    \centering
    \includegraphics[width=0.82\linewidth]{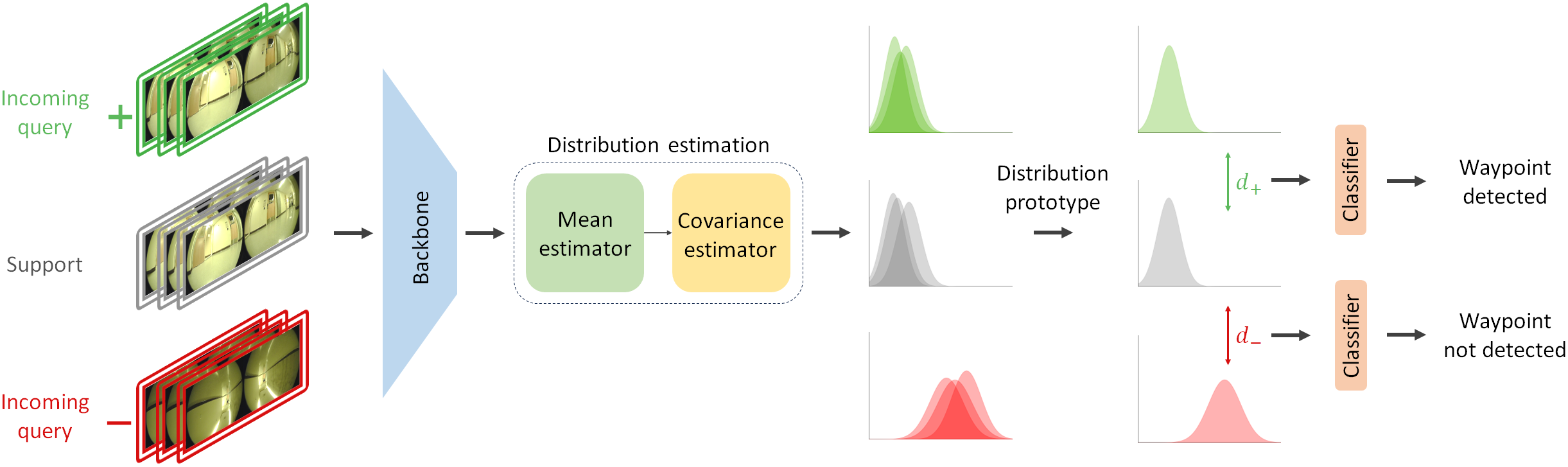}
    \caption{For each waypoint, a sequence of frames from an existing recording at the waypoint location (gray outline) are passed through the backbone and the distribution estimation model to obtain a corresponding class distribution for each frame (shown in gray). The distributions are then combined to form the waypoint prototype, which will be stored in memory for use during wayfinding. For a query frame sequence, the distribution is similarly obtained. The distance between the memory prototype and the real-time calculated query distribution is obtained and fed to the classifier for detection (shown in green and red respectively for a match and a mismatch). Once a waypoint is detected, the corresponding navigation action is retrieved from the LUT to condition the low-level continuous control module.}
    \label{fig:model}
\end{figure*}

Fig.~\ref{fig:model} shows the waypoint detection model. In order to train the proposed technique for use in few-shot navigation, we collect and use a dataset of various courses, with multiple repetitions (or laps) of each. For each lap of each course, the frames are split into segments corresponding to positive or negative examples for each waypoint within that course. Frames at which the high-level navigation action corresponding to the $n^\text{th}$ waypoint can be safely initiated will be used as positive class examples for that waypoint. All the frames beyond the positive segment of the $(n-1)^\text{th}$ waypoint and prior to the $n^\text{th}$ waypoint's positive segment will be used as the negative class examples for the $n^\textit{th}$ waypoint.

In each classification task during training, we construct the support distribution for the positive class (i.e. a given waypoint) as follows: first, we choose a random lap from a random course, then we make a random selection of $s$ consecutive images from a random waypoint in the lap. We then pass each image through the backbone and present to mean and covariance modules to retrieve individual distributions and obtain a combined prototypical estimate of the distribution associated with the waypoint. We then obtain query distributions by sampling $s$ consecutive frames from either the positive or negative classes. $q_p$ positive query distributions are constructed from the positive segment associated with the same waypoint but in different recordings of the path. $q_n$ negative distributions are taken from the corresponding negative segment in either the same lap or different laps. Combined distributions are then created in a similar fashion as the positive support distribution.

Next, the distance between each of the $q_p+q_n$ queries and the positive support distribution is calculated and presented to a classifier module. The classifier estimates the probability of the query matching the waypoint. Binary cross-entropy loss is then calculated between true and predicted waypoint assignments for queries to update model parameters.

At inference, the vehicle memory is populated with one or multiple combined distributions per waypoint, corresponding to consecutive frames from a single previous recording of waypoint locations. Similar to training time, query distributions are formed by combining a number $n_q$ of the most recent consecutive frames captured via the vehicle camera. The distance between the incoming query and the memory distribution associated with the upcoming waypoint is calculated and presented to the classifier module to determine whether the waypoint is reached. If the output probability is larger than a threshold value, the waypoint is detected. In the case of multiple memory distributions for a waypoint, the maximum probability is considered. Once a decision is made on the executable action, it is sent to the maneuver control unit to update the network condition and accordingly change the throttle and steering response.

The maneuver control unit consists of a backbone that processes the input images and concatenates the resulting representation with a one-hot vector that represents the distinct high-level navigation action associated with the latest detected waypoint, for instance a right turn. As such, the dimensionality of the one-hot vector matches the number of permissible high-level action commands. This concatenation serves to condition the input to the regression module so that it produces control signals that match the desired action. The same action is maintained until a new waypoint is detected.

\section{Experiments}\label{sec:exp}

\subsection{Dataset and training}
The dataset consists of 36 courses, recorded in 11 University of California, Davis buildings. Of 36 courses, 18 are collected on a clockwise and 18 on a counterclockwise loop, with CW/CCW pairs covering the same closed paths. The recordings are repeated multiple times so that there are between 2 and 8 completed recordings for each course.

The recordings are done using a remote-controlled vehicle equipped with cameras. It is equipped with Arduino Mega for analog control of steering and throttle, as well as an onboard ZOTAC mini PC with 8 GB RAM, Nvidia GeForce RTX 2070 Mobile GPU, and Intel Core i7-9750H processor to run custom navigation models. Two ELP fisheye cameras with a 180$^{\circ}$ field of view are mounted on the car, each turned 30$^{\circ}$ outwards, creating a 60$^{\circ}$ relative angle between the two camera orientations. The car can be set to manual or autonomous, which we will use for online evaluation.

To segment each course into various landmark/non-landmark segments after collecting data, we use the output steering signal. A fixed number of frames before the onset or completion of a turn are generally considered to be positive samples for waypoint detection, i.e. any one of these frames would have been a safe candidate for initiating the waypoint action at the time of recording.

To train the model, 12 CW/CCW pairs are used. 6 courses are used for validation in order to identify the best-performing model. The remaining courses are reserved for testing. Since our main focus in this work is on landmark detection and given the limited data-collection budget, the data for training the maneuver control module included the test locations in the form of repeats of each road segment with random actions at various intersections. The corresponding high-level navigation action is provided as a condition to the control module during its training. We note that while the downstream low-level maneuver control module has been exposed to test courses, the course remains unseen for the waypoint detection model, and successful navigation hinges on successful landmark detection.

We use the ResNet-50 architecture~\cite{he2016deep} as the waypoint detection model backbone. We estimate the sample class mean through a linear layer applied to backbone-generated features. The covariance module and the classifier are fully connected modules with 2 and 4 layers, respectively. To ensure positive variances, Softplus activation is used at the covariance output. Sigmoid is applied at the classifier output, which denotes the probability that a waypoint match is detected given the distance input.

For initializing the ResNet-50 encoder, we use weights obtained through pretraining on ImageNet~\cite{russakovsky2015imagenet} in the self-supervised manner described in~\cite{caron2020unsupervised}, which is made available by the authors. We train the model in two phases. In phase 1, we freeze the backbone and only train the mean and covariance modules, as well as the classifier. This prevents dramatic unfavorable updates to the backbone as a result of random initialization of other components. Next, we pick the best-performing model from this phase and jointly fine-tune all model parameters. For training the control unit, we use an EfficientNet-B0 backbone~\cite{tan2019efficientnet} and concatenate the high-level navigation condition with the output before passing to a fully connected segment whose output will be used as a signal to steer the vehicle. For simplification, we fix the throttle and only consider steering. More details on training and hyperparameters are included in the Appendix.

\subsection{Offline evaluation}

We use the 6 unseen test courses to evaluate the performance of the waypoint detection model offline. For evaluation on each course, we construct several positive class distributions per waypoint using samples from a single lap and store them in memory. We then provide frames from other laps in the same course in sequence and construct query distributions by combining the most recent consecutive frames. The distance between the query and memory distributions from the upcoming waypoint is then provided to the classifier for detection, where the maximum match probability among memory distributions is considered. The accuracy is evaluated by comparing the ground truth and the predicted labels of each road segment.

\subsection{Online evaluation}

For online evaluation, the trained model is loaded onto the vehicle computer. Test location waypoint images extracted from a single lap recording (memory lap) and their corresponding high-level navigation actions are provided to the model to construct memory distributions and the LUT. It is then set to autonomous driving mode to navigate the path. Each course test is repeated four times in total, twice for each of the two different memory laps. The course-level (completed course) as well as the waypoint-level (correctly identified waypoints) success rates are reported.

\subsection{Results and ablation study}
\subsubsection{Offline evaluation results and the effect of backbone pretraining, metric, and image quality}
Table~\ref{tab:offline} shows offline evaluation results. It also demonstrates the effects of backbone pretraining and the metric used. Considering that in our proposed distribution embedding we impose uncorrelated features through a diagonal covariance, we also explore an alternate configuration where the input to the classifier consists of dissimilarities between univariate normal distributions. In the case of Euclidean distance, we equivalently consider univariate differences corresponding to the representation features. The reason for this is to also test for partial metric learning, where the choice of metric is made at the univariate level, but we give the classifier the added flexibility of learning how to recombine these univariate dissimilarities instead of hard coding it in the cumulative form.

In each case, the optimal threshold was picked based on performance on the validation set. As results demonstrate, backbone pretraining generally improves the results significantly. Further, in nearly all cases the incorporated self-supervised pretraining outperforms the supervised pre-trained counterpart, where pretraining uses ImageNet labels.

The proposed symmetrized Mahalanobis metric demonstrates the highest performance. However, Wasserstein and KL divergence-based methods perform worse than Euclidean when multivariate classifier input is considered; however, when aggregate univariate input is used they outperform the Euclidean-based implementation, pointing to a potential benefit in metric learning for this application.
\begin{table*}
    \centering
    \caption{Offline accuracy results (\%) and the effect of pretraining, measure of dissimilarity, and the effect of presenting difference as dissimilarity between multivariate representations or as aggregated dissimilarities between univariate representations corresponding to individual latent space dimensions.}
    \begin{tabular}{lcccccc}
        \toprule
         & \multicolumn{2}{c}{No Backbone Pretraining} & \multicolumn{2}{c}{Supervised Backbone Pretraining} & \multicolumn{2}{c}{Self-Supervised Backbone Pretraining} \\
        \cmidrule(lr){2-3} \cmidrule(lr){4-5} \cmidrule(lr){6-7}
       Dissimilarity & Multivariate & Aggregate Univariate & Multivariate & Aggregate Univariate & Multivariate & Aggregate Univariate \\
        \midrule
        Euclidean & 52.6 & 76.7 & 83.0 & 88.4 & 89.4 & 86.1 \\
        KL Divergence & 81.1 & 76.7 & 79.9 & 90.6 & 80.7 & 92.2 \\
        2-Wasserstein & 78.2 & 76.4 & 81.8 & 89.1 & 84.1 & 91.6 \\
        Sym. Mahalanobis & 56.4 & 65.2 & 84.9 & 92.6 & \textbf{92.8} & 91.1 \\
        \bottomrule
    \end{tabular}
    \label{tab:offline}
\end{table*}

We also evaluate the robustness of the model to changes in scenery and image quality. For scenery changes, we consider adding coarse dropout as a proxy for occlusion, as well as brightness changes. For image quality, we consider defocus blur and Gaussian noise. Each effect is evaluated at three different and controlled intensities (see Appendix). Table~\ref{tab:robustness} documents the results, which show that the model is quite robust to changes in the scenery within a relatively permissive range; however, it quickly fails with image quality corruptions. We attribute this to the specific transforms incorporated in training that exposed the model to similar scenery changes, but not to quality corruptions, which implies a potential improvement through training-time corruptions.
\begin{table}
    \centering
    \caption{The effect of scenery change and quality deterioration during test time on waypoint detection accuracy.}
    \begin{tabular}{lccc}
        \toprule
        Modification & Low & Moderate & Severe \\
        \midrule
        Coarse Dropout & 91.3 & 87.0 & 80.2 \\
        Brightness Change & 92.2 & 91.2 & 75.4 \\
        Defocus Blur & 49.9 & 49.9 & 49.9 \\
        Gaussian Noise & 49.9 & 49.9 & 49.9 \\
        \bottomrule
    \end{tabular}
    \label{tab:robustness}
\end{table}

\subsubsection{Online evaluation results}
In online evaluation, each of the 6 courses consists of 8 waypoints. As mentioned before, each course is evaluated 4 times, corresponding to two repetitions for each of the two memory laps. We also consider a separate, long course consisting of 20 waypoints and similarly test on it. Results are shown in Table~\ref{tab:onlinetests}, where the first 6 courses are denoted as CW and CCW navigation of Location IDs 1-3 and the long course with location ID 4. In all failed waypoint detection cases with the threshold of 0.65 picked based on offline validation set performance, while the probability at the waypoint location was high, it did not quite exceed the threshold. We subsequently conducted a limited online test with the default threshold of 0.5 where we tested on each route once. With this adjustment, no failures occurred in online tests.
\begin{table}
    \centering
    \caption{Online evaluation results. Each value denotes (course-level success rate, waypoint-level success rate).}
    \begin{tabular}{lcccc}
        \toprule
        & \multicolumn{2}{c}{Threshold = 0.65} & \multicolumn{2}{c}{Threshold = 0.5} \\
        \cmidrule(lr){2-3} \cmidrule(lr){4-5}
        Location ID & CW & CCW & CW & CCW \\
        \midrule
        1 & (3/4, 31/32) & (4/4, 32/32) & (1/1, 8/8) & (1/1, 8/8) \\
        2 & (3/4, 31/32) & (4/4, 32/32) & (1/1, 8/8) & (1/1, 8/8) \\
        3 & (4/4, 32/32) & (3/4, 31/32) & (1/1, 8/8) & (1/1, 8/8) \\
        4 & \multicolumn{2}{c}{(2/4, 77/80)} & \multicolumn{2}{c}{(1/1, 20/20)} \\
        \midrule
        Total & \multicolumn{2}{c}{(23/28, 266/272)} & \multicolumn{2}{c}{(7/7, 68/68)} \\
        \bottomrule
    \end{tabular}
    \label{tab:onlinetests}
\end{table}

\section{Discussion and future work}\label{sec:conclusion}

In this work, we proposed a two-stage, end-to-end technique for autonomous navigation consisting of a high-level waypoint detector based on few-shot learning, as well as a low-level maneuver control unit that controls the vehicle conditioned on the high-level navigation input. This technique only requires a minimal amount of data from critical waypoints on an unseen route and has a low memory and computation demand. We believe the demonstrated offline and online results serve as proof of concept and motivate further developments with the aim of significantly reducing the need for positioning devices, expensive repeated training, and extensive data from target environments.

We are conducting additional research to gauge the approach, its robustness, strengths and limitations, as well as efficient ways to improve it in more diverse environments including outdoor settings. Utilizing public road data (e.g. Google Street View) as a low-cost source of waypoint support examples constitutes another aspect of our continued work. Besides more realistic autonomous driving implementations, for instance, on a full-scale vehicle that uses street images paired with a more sophisticated maneuver control mechanism, several exciting directions remain unexplored. One is the regime in which the car has missed or misidentified a waypoint, which will likely throw the vehicle entirely off-path. To address this, limited additional and ideally native sensors such as the vehicle odometer and/or a compass can be incorporated, which can also make it easier to correctly identify waypoints and execute navigation actions in the first place. Depending on the context, an offline local map can also be used to reroute the mobile robot or vehicle to its most recent known location. Alternatively, a new online navigation task from the new location to the destination can be set up in the same manner as before. Another direction of focus is improving robustness to outdoor landscape changes over time for instance due to seasonal variations. The increasingly realistic synthetic data and state-of-the-art generative models will likely play a crucial role in this regard. Addressing these problems without substantial data collection, computation, and hardware overhead will be topics of particular interest.

\appendices
\section{Waypoint detection training and evaluation details}
At each frame, the left/right camera images are separately processed. They are resized to $224\times 224$ and normalized. During training, random rotation, color jitter, and coarse dropout are applied to the images. After obtaining the distances associated with each camera, they are concatenated and presented to the classifier. 

At each waypoint, 15 frames leading to the steering initialization/completion are regarded as positive frames for that waypoint, corresponding to a conservative viable range to start steering earlier than where the steering was initiated when recording. The number of consecutive frames $s$ combined into a single distribution is set to 10 for both support and query. The same number is also used in evaluation. During training, each training episode has 1 positive and 6 negative queries ($q_p$ and $q_n$, respectively). Parameters are updated after processing each episode batch size of 36 in phase 1 and 3 in phase 2. A total of 240 and 4000 iterations are processed in phases 1 and 2, respectively. Adam optimizer is used with an initial learning rate of $10^{-4}$ in phase 1 and $10^{-5}$ in phase 2. The learning rate is divided by two after 160 iterations in phase 1 and every 1000 iterations in phase 2. In phase 1, the model is evaluated on the validation set every 32 iterations as well as at the end. It is evaluated every 200 iterations in phase 2.

For offline testing of a course, waypoint frames of a single lap are processed to obtain support distributions associated with waypoint locations and store them in memory. A sliding 10-frame window across a 15-frame range yields 6 different combined memory distributions per waypoint. Once the memory is populated, we run through each test lap frame by frame, each time comparing the most recent combined distribution with each of the 6 memory distributions, and obtain the match probability as the maximum of the 6 resulting probabilities. A waypoint is detected if the maximum of the 6 probabilities is above a cutoff probability. Once a waypoint is traversed, the expected upcoming waypoint is updated and a new binary classification task is set up. The evaluation is repeated for all configurations of memory/test laps and average performance is reported.

Waypoint-level accuracy is calculated as follows: If a high-probability frame occurs in a non-waypoint segment, it will be counted as a false positive, and if no high-probability frames are detected in a waypoint segment, a false negative will be recorded. Each waypoint range is stretched by a fixed small number of buffer frames before and after the original 15-frame range to improve the metric. The frames prior to a waypoint account for the common occurrence of rising probabilities when nearing a waypoint and hence, erroneously amplifying the false-positive rate. The frames after the nominal waypoint frames approximate positions where a turn could still be safely initiated although a bit later than in the recording. As such, the offline accuracy presents an underestimation of performance. Per our observations, nearly all false positives correspond to early detections which often pose no challenge in practice as implied by the large gap between online and offline evaluation results.

For online evaluation, besides populating the memory, a lookup table is also stored with keys corresponding to waypoint IDs. Instead of feeding consecutive test frames offline, the car is set to autonomous to navigate the course. Both through the steering module training procedure and observing vehicle detections during the test, we ensure that successful performance at each waypoint follows a valid detection. If the car fails at a waypoint, it is set up immediately after that waypoint, and the upcoming waypoint ID is manually updated to evaluate the remainder of the course.

\section{Low-level maneuver control module details}
Data for the low-level maneuver control module is collected by running the car in test locations. At waypoint positions, each maneuver action (straight/left turn/right turn) is repeated 3-5 times. This data is added to the previously collected test location data to sensitize the model to the high-level action condition and avoid memorizing a specific navigation action at waypoints. We emphasize that data collection from the test location is not an inherent necessity nor limitation of the technique; rather, this decision was made given our limited data collection budget and given the nature of our collected dataset for, and primary focus on, waypoint detection.

The model is composed of a pretrained EfficientNet-B0 backbone, whose 1280-dimensional output passes through a linear layer and ReLU activation. At this stage, the 500-dimensional output is concatenated with a 3-dimensional one-hot vector denoting the action condition (left turn/right turn/straight), which then passes through two more layers with ReLU and sigmoid activations and output dimensions of 100 and 1, respectively. The output is considered as steering.

The model is trained using mean square error (MSE) loss between true and predicted steering. The two camera images are concatenated, resized to $104\times 224$, and normalized. During training, random color jitter, Gaussian blur, and horizontal flip are applied to the image. In the case of horizontal flipping, the steering is mirrored and in case the action condition is turning, it is adjusted to reflect a turn in the opposite direction. The model is trained for 100 epochs with Adam optimizer and a fixed learning rate of $10^{-4}$.

\section{Image corruption details}
Below we list the details of the applied corruptions for robustness tests. In each case, the corruption was applied to all test images. Defocus blur is applied exactly the same to all images for each intensity level, but brightness change, coarse dropout, and Gaussian noise were randomly applied with fixed permitted parameter ranges.
\begin{itemize}
    \item Coarse dropout (low/moderate/severe): max. number of holes: 3/3/3; max. box [height,width] (pixels): [40,40]/[80,80]/[120,120]
    \item Brightness change (low/moderate/severe): multiplicative factor range [min,max]: [0.75,1.33]/[0.5,2]/[0.33,3]
    \item Defocus blur (low/moderate/severe): blur kernel radius (pixels): 3/5/10; range for alias blur of defocusing: 0.1/0.3/0.5
    \item Gaussian noise (low/moderate/severe): noise standard deviation: 10/50/200
\end{itemize}

\section*{Acknowledgment}

The authors would like to thank Aryan Mondkar, Dechen Gao, and Mohamed Shais Khan for their contributions, including their help with data collection.

\ifCLASSOPTIONcaptionsoff
  \newpage
\fi

\end{document}